# Clinical Communication Processing with Models Trained on LLM-Generated Synthetic Data: A Structured Survey and Novel Application Case Studies

**Alexander Apartsin**[1] **Yehudit Aperstein**[2]

[1] School of Computer Science, Faculty of Sciences, Holon Institute of Technology (HIT), Holon, Israel

[2] Intelligent Systems, Afeka Academic College of Engineering, Tel-Aviv, Israel

**Abstract**

A great deal of clinical value is conveyed not through structured records but through communication: the spoken and written exchanges in which patients describe symptoms, clinicians reason and give instructions, ambulances hand over to emergency departments, and nurses pass on a shift. Such language differs from tabular data because meaning depends on speaker role, intent, causality, uncertainty, omission, and channel noise, and healthcare natural language processing (NLP) must interpret information as it is conveyed rather than as it is coded. Learning to do so requires large, diverse, and well-annotated corpora of clinical communication, which are scarce: authentic exchanges are private, fragmented across channels, and costly to annotate. Large language models offer a way forward, because they can transform structured or documentary clinical sources, such as records, diagnostic labels, symptom lists, or care plans, into written and transcribed spoken communication for downstream models.

We present a structured narrative survey of this emerging area, organized by source representation, communication form and participants, generation method, and downstream task, and complemented by thirteen novel application case studies that build clinical NLP systems for communication channels and languages with no labeled real-world data, spanning EMS pre-arrival reports, field-radio casualty documentation, nurse handoffs, patient-portal triage, and low-resource discharge communication, and showing that synthetic communication can bootstrap systems that could not otherwise be trained. Recurring observations include the competitiveness of fine-tuned encoder models over evaluated zero-shot baselines and the value of deliberately degraded communication for robustness. The central limitation is that most studies evaluate on held-out synthetic communication, with train-on-synthetic, test-on-authentic evidence still accumulating. We conclude that synthetic clinical communication is becoming a practical research resource; establishing it as reusable clinical infrastructure will build on authentic-data transfer together with attention to safety, privacy, and external validation.

## 1 Introduction

Clinical care runs on communication. Before any note is filed, a patient describes symptoms, a paramedic radios a pre-arrival report, a nurse hands off a shift, a dispatcher triages an emergency call, and a clinician gives follow-up instructions. These exchanges carry the reasoning, urgency, and intent that downstream systems most need, yet they are captured (when captured at all) as noisy, unstructured, privacy-sensitive text. As data-hungry machine learning methods have matured, so has the demand for large, diverse, annotated corpora of clinical communication.

The stakes are high, because much of the clinically decisive information in medicine never reaches a structured field. The patient history alone yields the correct diagnosis in roughly four out of five cases[84], and a large share of the electronic health record is free text whose notes materially improve prediction and phenotyping[88]. This communicative layer is also fragile: communication failures are among the leading contributors to sentinel events[86] and were implicated in thirty percent of malpractice cases, 1,744 deaths, and 1.7 billion dollars in cost over five years[85]. Information is routinely lost at paramedic-to-emergency-department handover[87] and even in the language of an emergency call, where delayed recognition of cardiac arrest lowers survival[90]. As patient-portal messaging surges and adds to documentation burden[89], the value carried in clinical communication is both larger and more at risk than the structured record suggests.

Collecting and sharing such corpora remains difficult. Clinical communication is protected health information; de-identification is imperfect and costly; institutional data governance constrains access and reuse; expert annotation is slow and expensive; and clinically important phenomena such as rare diseases and adverse events are, by definition, underrepresented. These barriers slow progress and limit the reproducibility of clinical NLP research.

The emergence of instruction-following LLMs has changed what is possible. Given a scenario, a schema, or a few examples, an LLM can render clinically plausible narratives, conversations, messages, and reports on demand, optionally carrying their own annotations. Importantly, the source defines the intended facts and the generator realizes them in language; the model does not invent clinical truth. This capability turns synthetic clinical communication from a privacy fallback into a controllable resource for building and evaluating healthcare NLP.

**Synthetic clinical communication** is artificially generated natural-language content representing written or spoken exchanges among participants in healthcare: patient narratives, dialogues, messages, instructions, handoffs, emergency reports, and transcribed radio or telephone communication in which clinically relevant information is conveyed through contextual, informal, incomplete, uncertain, or noisy language. The clinical content used to generate it may originate from structured records, diagnostic labels, symptom lists, notes, care plans, guidelines, or manually constructed scenarios; these are generation inputs, not the downstream textual modality studied here. Inclusion is therefore determined by the generated language and its downstream use, not by the format of the source. The organizing pipeline is accordingly *clinical source → LLM-generated communication → downstream healthcare NLP model*. Our object of study is that final stage: NLP systems that process clinical communication and are trained, wholly or in part, on such synthetic data. The word *synthetic* qualifies the training data, not the real communication these systems are ultimately built to serve.

Within this scope we include patient–clinician conversations, doctor-to-patient instructions, telemedicine consultations, paramedic-to-hospital and emergency-dispatch traffic, nurse and clinician handoffs, and patient-portal messaging, whatever the source from which they are generated. We exclude work whose output is synthetic structured or tabular records, signals, or images; internal documentation with no communicative function; and studies in which an LLM only classifies existing text without generating communication. A discharge summary, for instance, is in scope when it functions as communication to the patient or the next provider, and out of scope when treated purely as an archival record.

This paper is a structured narrative survey complemented by application case studies, organized to answer seven questions: (RQ1) which forms of written and transcribed spoken clinical communication are generated with LLMs; (RQ2) which clinical source representations are transformed into them; (RQ3) which communication properties, such as role, intent, uncertainty, missingness, style, and noise, are controlled during generation; (RQ4) which downstream tasks use the result for training, augmentation, or evaluation; (RQ5) how clinical fidelity, communication realism, label quality, privacy, and downstream utility are evaluated; (RQ6) what evidence exists that models trained on synthetic communication generalize to authentic communication; and (RQ7) which reusable design patterns and failure modes recur across the case studies. Our contributions are to organize a fragmented literature around the source-to-communication pipeline; to survey recent generation and evaluation methods; to present thirteen novel application case studies that build clinical NLP for communication channels and languages lacking any labeled real-world data, demonstrating feasibility where models could not otherwise be trained; and to distill reusable design patterns and open problems in an emergent field.

Existing reviews cluster into two camps that leave our target uncovered. Broad "LLMs in medicine" surveys treat text generation as one application among many[23,24], while synthetic-clinical-text reviews pool clinical notes, tabular EHR, and time series under generic utility and privacy lenses[19,20,21,22,36]. Even the dedicated survey of medical-dialogue generation frames generated dialogue as a conversational system's output rather than as reusable synthetic training data[25,26], and stops at the patient-provider chat channel. No prior review organizes the field around clinical communication as a modality spanning patient–clinician dialogue and ambient scribing[28,33,34,35], doctor-to-patient instructions, telemedicine, EMS and SBAR handoff, emergency dispatch[38], and patient-portal messaging, nor pairs that taxonomy with concrete application studies and a channel-specific privacy analysis[31]. This paper fills that gap: it unifies these fragmented channels, maps which datasets are real versus LLM-synthetic, and shows through application studies how synthetic clinical communication trains and evaluates downstream NLP where real conversational data is scarce, restricted, or absent, notably the under-served EMS, dispatch, and handoff channels. The closest prior work is a typology of synthetic clinical dialogue datasets[83], but it organizes by generation method and treats dialogue as a single undifferentiated category, without the channel structure, the under-covered EMS, dispatch, handoff, and portal channels, or the paired application studies offered here.

## 2 Literature Review

This section reviews the field in four parts: the modalities of clinical communication and the tasks they support (2.1); how large language models generate synthetic clinical communication (2.2); how such data is evaluated (2.3); and what the literature reports, organized by downstream task (2.4).

Efforts to unlock the value in clinical communication predate synthetic data. Ambient systems that turn a recorded consultation into a note, together with their benchmarks and shared tasks[27,33,34,58], are now moving into practice as clinician-facing scribes[93]. Information extraction has been applied to paramedic narratives[94] and, in emergency medicine more broadly, is the subject of dedicated reviews[95,96]; machine learning on the audio and language of emergency calls has improved dispatcher recognition of cardiac arrest in a randomized trial[91,92]; and patient-portal messages are increasingly triaged and routed automatically[97]. What these lines share is a dependence on scarce, privacy-restricted real communication for training and evaluation, the bottleneck that synthetic generation, the subject of this survey, sets out to relieve.

### 2.1 Clinical Communication in Healthcare AI

#### *2.1.1 Modalities of Clinical Communication*

Clinical communication comprises a family of channels, each defined by who is speaking to whom, under what time pressure, and through what medium, and each with its own characteristic way. A patient–clinician conversation is a two-party, turn-taking exchange in which the clinician steers while the patient supplies symptoms, often in lay, emotional, or incomplete terms. Doctor-to-patient instructions run the other way: discharge summaries, follow-up plans, and medication guidance are largely one-directional and are addressed as much to the next provider as to the patient. Telemedicine consultations move these encounters onto remote text or voice, adding transcription and connectivity artifacts. Paramedic-to-hospital communication and emergency dispatch are terse, urgent, and spoken over noise, burdened further by automatic-speech-recognition errors and, in the field, by radio dropouts. Nurse and clinician handoffs, epitomized by the SBAR format, are structured precisely because omissions at transfer are a leading source of harm. Patient-portal messaging, finally, is asynchronous and patient-initiated, arriving in short, informal, error-strewn bursts. Table 1 sets these channels against the dimensions that matter for generation and modeling.

***Table 1.*** *Modalities of clinical communication and their defining properties; the final column links each channel to the application studies in Section 3.*

| Channel | Participants | Medium | Register / urgency | Characteristic difficulty | Studies |
|---|---|---|---|---|---|
| Patient–clinician conversation | patient, clinician | spoken or typed dialogue | exploratory, moderate | lay, incomplete self-description | 3.1 |
| Doctor-to-patient instructions | clinician to patient / next provider | written | directive, low | implicit timing, cross-lingual gaps | 3.2 |
| Telemedicine consultation | patient, remote clinician | remote text or voice | exploratory, moderate | transcription, connectivity artifacts | 3.3 |
| Patient-portal messaging | patient to care team | asynchronous text | informal, variable | noise, typos, ambiguity | 3.3 |
| Paramedic-to-hospital (EMS) | paramedic, emergency department | spoken / radio | terse, high | ASR noise, incompleteness | 3.4 |
| Emergency dispatch / field radio | caller or field, dispatcher | spoken / radio | terse, critical | dropouts, chaos, slang | 3.4 |
| Nurse / clinician handoff (SBAR) | outgoing, incoming staff | structured spoken / written | structured, high | completeness at transfer | 3.4 |

Real reference corpora exist for a handful of these channels, but each carries limitations that motivate synthetic generation. Large doctor–patient dialogue collections such as MedDialog[27] provide scale, yet they consist mostly of online consultations in a few languages and carry no task labels, so they cannot directly supply triage severities, decision spans, or the other targets a downstream model must learn. Ambient-scribing benchmarks that pair consultations with notes, including ACI-Bench[33], MTS-Dialog[34], and the mock consultations of PriMock57[35], are carefully annotated but small, from tens to a few hundred encounters, and confined to primary-care visits, so they cover neither rare presentations nor other channels. Note-conditioned synthetic dialogue such as NoteChat[28] begins to relax the size constraint but inherits the source notes' distribution. Crucially, none of these resources covers the EMS pre-arrival, emergency-dispatch, SBAR-handoff, or non-English portal channels at all, and every real collection remains privacy-restricted, costly to de-identify, and thin in the long tail. Synthetic generation targets exactly these gaps: the channels, languages, labels, and rare cases that real corpora do not supply.

### 2.1.2 Knowledge and Intent Embedded in Communication

Beyond named entities, clinical communication encodes the reasoning, causality, urgency, uncertainty, and intent that make individual facts actionable. A single exchange may carry symptoms and patient self-descriptions, diagnoses, medications, and procedures; the urgency or severity that sets a message's triage priority; decisions and instructions with timing, such as when to start, stop, or continue a treatment and when to follow up; and, at handoff, the question of completeness, what is stated versus what is missing or under-specified. It also carries temporal structure, emotion and distress, and patient concerns, all wrapped in the reliability and noise of real channels: transcription errors, interruptions, code-switching, and slang. Extracting any of these is a distinct clinical NLP task. What makes them hard is that communication conveys more than facts; the same clinical event surfaces very differently depending on who says it and how. The structured order “stop warfarin,” for instance, may arrive as “don't take your blood thinner tonight,” forcing a model to resolve the reference, the timing, the speaker, and whether the instruction is active. Table 2 catalogs the recurring communication phenomena and the downstream difficulty each creates.

***Table 2.*** *Communication-specific phenomena and the downstream difficulty they create.*

| Phenomenon | Example | Downstream difficulty |
|---|---|---|
| Implicit reference | “the small white pill” | concept resolution |
| Missing information | no allergy status mentioned | completeness analysis |
| Causal explanation | “I stopped because it made me dizzy” | relation extraction |
| Temporal ambiguity | “since last week” | temporal normalization |

| Self-correction | "two tablets, actually one" | robust extraction |
|---|---|---|
| Lay expression | "my heart skips" | concept mapping |
| Distress | fragmented, urgent message | triage and prioritization |
| Role dependence | patient versus paramedic wording | interpretation |
| Code-switching | Hebrew text with English drug names | multilingual NLP |
| Transcription (ASR) error | distorted drug name | noise robustness |

### 2.1.3 Clinical NLP Tasks

The tasks in Table 3 recur throughout the survey and motivate the downstream-utility view of synthetic clinical communication.

***Table 3.*** *Core clinical NLP tasks addressed by synthetic clinical communication.*

| Task | Objective | Typical output |
|---|---|---|
| Named entity recognition | Locate clinical concepts | Typed text spans |
| Relation extraction | Link entities | Entity pairs / graphs |
| Temporal extraction | Order events in time | Timelines |
| Information extraction | Populate structured fields | Records / tables |
| Summarization | Condense the record | Abstractive summaries |
| Clinical coding | Assign standard codes | ICD / SNOMED codes |
| Question answering | Answer clinical queries | Spans / free text |
| Decision support | Recommend / flag | Alerts / suggestions |

## 2.2 Generating Synthetic Clinical Communication

### 2.2.1 Why Generate Synthetic Clinical Communication?

Synthetic generation addresses several bottlenecks at once, and different motivations lead to different design priorities. Foremost is privacy: text with no real patient behind it can be shared, published, and reused without the consent and de-identification burdens that lock away real conversations. Generation can also reduce and redistribute the annotation burden, because provisional labels can be derived from the source scenario rather than added by expensive expert review. It gives direct control over the long tail, so rare presentations and dangerous edge cases can be sampled deliberately rather than awaited, and underrepresented classes can be up-generated to correct imbalance. Finally, it enables controlled experimentation and benchmark construction: versioned, reproducible evaluation sets in which difficulty and distribution are set by design rather than confounded by whatever real data happened to be available.

### 2.2.2 Generation Approaches

Approaches to generating synthetic clinical communication span a spectrum from lightly conditioning a single model to orchestrating many, and they are usually combined rather than used in isolation. The simplest is *prompt engineering*: instruction and few-shot conditioning, sometimes enriched with domain knowledge injected into the prompt[29]. It is fast and flexible, but its output is brittle to prompt wording and prone to stereotyped phrasing. *Scenario-driven generation* samples from an explicit space of patients and encounters, so that coverage is controlled by design rather than left to the model's priors; the cost is the effort of building and validating that scenario space. *Template-guided generation* constrains structure with genre scaffolds such as SOAP or SBAR, which guarantees well-formed documents but can render them formulaic when the template dominates the content. *Multi-agent simulation* assigns separate patient and clinician agents that converse[28,60,62,72,73], producing realistic multi-turn and multi-party dialogue; the risk is that agents drift, collude toward easy cases, or contradict a fixed ground truth. *Self-refinement* adds critique-and-revise loops in which the model or a panel of critics improves realism and consistency, though a critic that shares the generator's blind spots may simply certify its own errors. *Retrieval-augmented generation* grounds the text in guidelines, ontologies, or exemplars[74], improving factuality while importing whatever biases and gaps the retrieval corpus contains. Distinct from all of these is *data augmentation*, which does not invent encounters but transforms existing text through paraphrasing, back-translation, controlled noise injection, or rewriting a clean record into graded noise variants[13,70]. Augmentation is cheap and preserves gold labels, and deliberately degraded variants can even improve robustness, as our EMS study shows (Section 3.4); its limitation is that it can only vary what the seed data already contains, and so does little for genuinely unseen presentations. In practice a pipeline chains several of these directions, for instance scenario sampling to fix the case, template scaffolding to shape the document, multi-agent play to voice it, and a judge to filter the result.

### 2.2.3 Automatic Annotation

A distinctive advantage of generative pipelines is that labels can be produced with the text rather than after it. In the strongest form, embedded ground truth, the generator emits the text and its labels jointly, as when a triage severity or a target action is fixed in the prompt and the message written to match it. These label-first and label-after directions are the two poles of clinical data generation[37]. A weaker but more flexible alternative runs a second labeling pass, or a panel of model judges, over already-generated text, an auditing step now studied in its own right[75]. The scenario parameters used during generation can also be retained as structured metadata, and noisy programmatic signals can be aggregated as weak supervision at scale. Several of the application studies in Section 3 combine these: labels fixed at generation time, then audited by an independent model judge before a record is admitted.

#### 2.2.4 Dataset Construction

Turning a generator into a dataset is a design problem in its own right. A useful corpus needs diversity and coverage across scenarios, phrasing registers, and patient demographics; explicit difficulty control, with hard and ambiguous cases included on purpose rather than by accident; and balancing across classes together with deliberate long-tail sampling, so that rare but critical situations are represented. The application studies that follow make these choices concretely, for example by generating messages at several noise levels, across distinct writer profiles, or with channel reliability deliberately degraded.

### 2.3 Evaluating Synthetic Clinical Communication

Evaluation is the least standardized part of the field. We organize it into four levels, from surface text quality to downstream utility, with the concrete metrics used at each level collected in Table 4. Throughout Section 3 the headline downstream metric is macro-averaged F1, which weights every class equally and is therefore the right summary for the imbalanced, safety-critical classes typical of clinical communication.

***Table 4.*** *Four levels for evaluating synthetic clinical communication, with the questions they ask and the metrics typically used.*

| Level | Question | Typical metrics |
|---|---|---|
| Text quality | Is the language fluent and coherent? | Perplexity, readability indices, human fluency and coherence ratings |
| Clinical quality | Are the facts correct and internally consistent? | Clinician rubric scores, factual-agreement and contradiction rate, hallucination and omission rate |
| Dataset quality | Is the corpus diverse, faithful, and safe? | Lexical and embedding diversity, distributional distance to real text, duplication rate, membership-inference and PHI-leakage tests |
| Downstream utility | Do models trained on it work? | Macro-F1, precision and recall, AUROC, and the train-on-synthetic / test-on-real (S→R) gap |

The **train-on-synthetic, test-on-real** protocol is the most decisive evidence of utility. It helps to label each study by its evaluation regime: $S{\to}S$ (train and test on synthetic data), $S{\to}R$ (train on synthetic, test on real), $R{+}S{\to}R$ (train on real plus synthetic, test on real), and $R{\to}R$ (a real-data baseline), with cross-generator ($S_1{\to}S_2$) and human-written ($S{\to}H$) variants as stronger tests. As Section 3 makes explicit, almost all of the application studies surveyed here operate in the $S{\to}S$ regime, which is why transfer to authentic communication remains the field's central open question.

Privacy deserves its own evaluation: de-identification alone can be insufficient, and synthetic notes that match real utility may inherit real privacy risk[31], which motivates dedicated privacy-and-utility metrics for medical synthetic data[32].

### 2.4 Existing Applications, by Task

We organize the surveyed literature by **downstream task** rather than by generation technique, because the task ultimately determines which generation and evaluation choices matter.

Several generic NLP problem settings, though not originally clinical, map directly onto clinical conversation and recur across our application studies: recognizing entities from implicit contextual cues in first-person narratives[18], a setting also studied for noisy user-generated and conversational text[39,40]; comparing text content through a question-answering lens[6], as in QA- and NLI-based consistency checking[41,42,43]; extracting critical information from noisy transcribed communications[5], paralleling slot-filling from air-traffic and other radio speech[44,45]; and measuring the efficiency of diagnostic questioning[1], a problem with a deep line in symptom- checking and information-seeking clinical dialogue[46,47,48,49]. Even aspect-based sentiment, developed as a controlled synthetic benchmark in the education domain[11], transfers to patient-message sentiment and distress. Each was first studied in an adjacent domain, yet all transfer cleanly to history-taking, handoff, dispatch, and triage; we treat them as reusable problem templates throughout this section.

Table 5 summarizes the surveyed tasks, the role synthetic communication plays in each, and representative work; the transferable settings noted above recur across several of them.

*Table 5. Downstream clinical NLP tasks supported by synthetic clinical communication. Bracketed entries [n] cite surveyed literature; § entries point to the application studies in Section 3.*

| Task | Role of synthetic communication | Representative references |
|---|---|---|
| Information extraction (entities, relations, time) | Emit gold spans and relations with the text; recover decisions, follow-ups, and casualty fields; distil small extractors from large teachers | [4], [5], [18], [39], [40], [44], [45], [71], [79] · §3.2.1, §3.4.2 |
| Summarization (dialogue-to-note, impression, discharge) | Pair synthetic dialogue or source with target summaries; judge factuality with QA-based metrics | [26], [28], [33], [34], [35], [41], [42], [57], [58], [59] |
| Clinical coding | Up-sample rare codes; rank models without a trustworthy coded benchmark | [8], [30] |
| Diagnostic reasoning and questioning | Interactive question selection under partial information; robustness to noisy narratives | [1], [2], [46], [47], [48], [49] · §3.1.1, §3.1.2 |
| Question answering | QA- and NLI-based consistency checking; response-generation strategies | [6], [7], [41], [42], [43] |
| Triage, routing, and decision support | Class-imbalanced patient messages and pre-hospital calls; portal and emergency triage | [3], [38], [64], [66], [67] · §3.1.3, §3.3.1–3.3.6, §3.4.1 |
| Handoff and completeness checking | Controlled omissions in SBAR and handover notes | [65] · §3.4.3 |
| Patient simulation and professional training | Virtual standardized patients; controllable learner and patient simulation | [10], [14], [15], [50], [51], [52], [53], [54], [55], [56], [60], [61], [72], [73] |
| Benchmark construction | Controlled, versioned synthetic benchmarks and shared tasks | [8], [9], [10], [11], [12], [16], [17], [68], [69] · §3.1.2 |
| Privacy, augmentation, distillation (cross-cutting) | Augment, distil, and audit synthetic corpora; assess memorization and differentially private generation | [29], [30], [31], [32], [70], [71], [77], [78], [79], [80], [81], [82] · §3.3.5 |

## 3 Application Studies

This section presents thirteen application case studies, grouped by communication channel, each of which transforms a clinical source into synthetic communication and trains or evaluates models on it. They are illustrative rather than exhaustive, selected to span the channels of Section 2 rather than through systematic sampling. Each is a novel contribution in its own right: it builds a working system for a channel or language where labeled real-world data is unavailable, making synthetic communication the enabling resource rather than a mere convenience. For every study we discuss the clinical value at stake, the source and why authentic data is unavailable, how the language model produced the communication, and how the resulting models compare; the exposition varies with the study, leaning on prose, a results table, or a figure as best fits the material. Almost every study evaluates in the $S \rightarrow S$ regime (Section 2.3), that is, on held-out synthetic communication, so transfer to authentic communication remains the central open validation step, which we take up in Section 4. Table 6 summarizes all thirteen studies.

*Table 6. Clinical-communication processing studies, by channel, using models trained on LLM-generated synthetic data. Generators are the LLMs used to synthesize the training text; results are the headline downstream metric.*

| # | Study (§) | Channel | Synthetic generator | Headline result |
|---|---|---|---|---|
| 1 | Adaptive diagnostic questioning (3.1.2) | Patient–clinician diagnostic conversation | GPT-4o-mini (SDPD-seeded cases) | Adaptive questioning benchmark, partial-information tiers |
| 2 | Diagnosis from noisy self-descriptions (3.1.1) | Patient self-description (noisy) | Llama-3.1-8B (noise rewrite) | FLAN-T5 87.1% acc under heavy noise |
| 3 | Urgency triage from complaints (3.1.3) | Patient arrival complaint | GPT-4 (free-text complaints) | DistilBERT 0.96 acc / 0.83 F1 |
| 4 | Clinical-decision extraction (3.2.1) | Hebrew discharge instructions | GPT-4o-mini (518 summaries) | Reranker strict-F1 0.4273; relaxed-F1 0.9256 |
| 5 | Administrative portal-message triage (3.3.1) | Patient-portal messaging | Qwen2.5-0.5B, local (2.8k) | DistilBERT 0.81 F1 vs 0.29 zero-shot |

| 6 | Clinical-priority portal triage (3.3.3) | Patient-portal messaging | GPT-4o (3k longitudinal) | ClinicalBERT + safety cascade 0.997 acc |
|---|---|---|---|---|
| 7 | Postpartum severity triage (3.3.2) | Postpartum patient messages | GPT-4o-mini (3k) | BioBERT cascade 0.975; GPT-4o-mini 0.98 |
| 8 | Oncology distress classification (3.3.4) | Oncology patient messages | Gemma2-27B, local (1.3k) | DistilBERT distress macro-F1 0.86 |
| 9 | Medication-question risk classification (3.3.5) | Patient medication questions | GPT-4.1 (critical-case augmentation) | BioBERT / BlueBERT 0.90 macro-F1 |
| 10 | Home-care status detection (3.3.6) | Home-care status messaging | GPT (symptom + vitals) | LightGBM + TF-IDF fusion 0.97 F1 |
| 11 | EMS report routing (3.4.1) | Paramedic-to-ED pre-arrival | GPT-4.1-mini (8.6k, + ASR noise) | BioClinicalBERT 0.38 macro-F1 (specialty) |
| 12 | Casualty-record reconstruction (3.4.2) | Field casualty radio to hospital | GPT-4o (500, + noise) | AlephBERT 79.3% EM / 0.80 F1 |
| 13 | SBAR completeness checking (3.4.3) | Nurse and clinician handoff | OpenAI batch (5k) | ClinicalBERT class-F1 0.75; hierarchical 0.79 |

## 3.1 Patient–Clinician Conversations and Self-Descriptions

In a patient–clinician conversation the patient supplies symptoms and the clinician reasons toward a diagnosis; processing this exchange matters because the decisive signal is carried in lay, often noisy language rather than in structured fields. Recent work simulates the full conversation with multi-agent, self-refining models[60,61,72] and generates synthetic dialogue from notes for ambient scribing[57,28]. The three studies below concentrate on the patient's side of that exchange: how a model can learn from, and reason over, the way patients actually speak.

### 3.1.1 Diagnosis from Noisy Patient Self-Descriptions

Patients rarely describe their symptoms in the tidy vocabulary of a textbook: they ramble, hedge, digress, and reach for lay terms, so an intake or triage model that understands only clean descriptions fails exactly where it is most needed. This study frames diagnosis as multi-class classification over twenty-four disease categories from free-text self-descriptions, and asks specifically how accuracy holds up as the language grows noisier. The nearest public resource, the 24-class Symptom2Disease set[105], is small and written in clean prose; naturally noisy patient narratives paired with confirmed diagnoses are neither public nor easily gathered under privacy constraints, so a graded-difficulty corpus cannot be assembled from real records. To build one, a local Llama-3.1-8B model rewrites 1,200 clean symptom descriptions into medium- and heavy-noise variants, introducing repetition, off-topic asides, and personal narrative while preserving the original diagnosis label, giving 2,400 cases at controlled noise levels. Four classifiers are then trained and compared as the language degrades: a TF-IDF Naïve Bayes baseline, BERT, ClinicalBERT, and a generative FLAN-T5. FLAN-T5 is the most noise-robust, degrading least under heavy noise where the Naïve Bayes baseline collapses (Table 7); the accompanying benchmark of LLMs on noisy patient narratives reports the same pattern[2].

***Table 7.*** *Diagnostic accuracy by narrative noise level (synthetic test).*

| Model | Clean | Medium noise | Heavy noise |
|---|---|---|---|
| Naïve Bayes (TF-IDF) | 93.8% | 79.2% | 77.5% |
| BERT | 98.3% | 86.7% | 79.2% |
| ClinicalBERT | 97.9% | 83.8% | 86.2% |
| FLAN-T5 | 97.1% | **92.5%** | **87.1%** |

### 3.1.2 Adaptive Diagnostic Questioning under Partial Information

Diagnosis in practice is interactive: a clinician asks a question, updates a differential, and asks again, and the skill that matters is choosing the question that resolves the most uncertainty. Most benchmarks bypass this by handing a model the complete case, so this study instead frames diagnosis as an interactive process in which an LLM clinician interrogates an LLM patient that holds only part of a case and commits to a diagnosis after each round. Interactive benchmarks such as MediQ[48] recast exam vignettes into partial-information questioning, but organic diagnostic conversations with ground-truth outcomes are rarely released, so the encounters are simulated: GPT-4o-mini instantiates cases from the Symptom-Disease Prediction Dataset (132 symptoms, 41 diseases) at three information-reveal tiers (Table 8) and plays the patient. Question selection follows an information-gain objective, choosing the next question $q$ that most reduces diagnostic uncertainty, $q^* = \operatorname{argmax}_q I(D; A_q \mid H)$, where $D$ is the diagnosis, $A_q$ the answer, and $H$ the dialogue history; each round's diagnosis is scored against the ground-truth prognosis. Because this is an evaluation benchmark rather than a trained classifier, its contribution is a controllable testbed: the reveal tiers make questioning efficiency measurable and anchor the diagnostic-questioning benchmark line[1].

***Table 8.*** *Information-reveal tiers used to probe diagnostic-questioning efficiency.*

| Tier | Symptoms revealed to patient agent | What it measures |
|---|---|---|
| Full | 100% | Ceiling accuracy with complete information |
| Partial | 80% | Robustness to mild information gaps |
| Sparse | 50% | Questioning efficiency under high uncertainty |

### 3.1.3 Urgency Triage from Arrival Complaints

When a patient reaches the emergency department, the first decision, made from a short account of the problem, is whether the case is urgent. This study frames that decision as binary text classification, urgent versus non-urgent, over first-person arrival complaints. Real triage data such as MIMIC-IV-ED[107] pairs a telegraphic chief-complaint field with a five-level acuity, but it is access-gated and lacks first-person prose, so GPT-4 generates natural, non-clinical complaint texts that imitate how patients describe their symptoms, with the original triage labels collapsed to a binary target. Three models are compared: a TF-IDF logistic-regression baseline, a fine-tuned DistilBERT encoder, and a fine-tuned T5 generator (Table 9). DistilBERT attains the best accuracy and F1, while the logistic-regression baseline retains the highest recall, a trade-off that is consequential for a safety filter deliberately biased toward over-triage.

***Table 9.*** *Binary emergency-department triage (synthetic test).*

| Model | Accuracy | F1 | Recall |
|---|---|---|---|
| TF-IDF + logistic regression | 0.810 | 0.653 | **0.873** |
| DistilBERT (fine-tuned) | **0.964** | **0.826** | 0.594 |
| T5 (fine-tuned) | 0.931 | 0.549 | 0.536 |

## 3.2 Doctor-to-Patient Instructions and Care Handover

Here the clinician communicates decisions to the patient and to the next provider, through discharge summaries and follow-up plans; processing this channel matters because those instructions must become scheduled, auditable actions rather than free prose. Discharge and follow-up communication has become an active target, with simulation-and-generation benchmarks[68] and shared tasks on discharge-document generation[69]. The study below extracts structured decisions from synthetic discharge summaries in a low-resource language.

### 3.2.1 Extracting Clinical Decisions from Hebrew Discharge Summaries

A discharge summary is the moment a hospital hands its decisions to the patient and to whoever provides care next, so turning that prose into a structured, auditable record, which drug was started or stopped, which procedure was performed or planned, is a clinically useful extraction task. This study formalizes it as recovering, from each Hebrew summary, a set of four-tuples of verbatim span, decision type, canonical concept, and action. The task has an English analogue, MedDec[106], which labels clinical decisions over MIMIC-III discharge summaries, but no Hebrew labeled corpus exists, and Hebrew clinical text is scarce, right-to-left, and heavily code-switched with English drug names, so there is nothing shareable to learn from in this language.

To create training data, GPT-4o-mini (Batch API) generates 518 synthetic Hebrew summaries carrying 1,222 gold spans, anchored to MedDec- and MIMIC-III-derived clinical targets by marked-span anchoring, with every span validated to appear verbatim in its summary so that the labels are recoverable by construction. Four systems are then compared: an XLM-R BIO tagger, an sklearn candidate reranker, a Gemini candidate selector, and a Qwen2.5-1.5B few-shot baseline (Table 10, Figure 1). The reranker is strongest under strict boundary matching (0.427) and Gemini under relaxed matching (0.926); the wide relaxed-minus-strict gap shows that errors are largely boundary misalignment rather than wrong regions. Because both training and test data are synthetic, transfer to genuine Hebrew notes remains the open question.

***Table 10.*** *Clinical-decision extraction from Hebrew discharge summaries: overall F1 (v2 test, 81 summaries).*

| Model | Strict F1 | Relaxed F1 |
|---|---|---|
| Qwen2.5-1.5B (few-shot) | 0.206 | 0.650 |
| XLM-R (BIO) | 0.326 | 0.829 |
| sklearn reranker | **0.427** | 0.677 |

| Gemini selector | 0.318 | **0.926** |
|---|---|---|

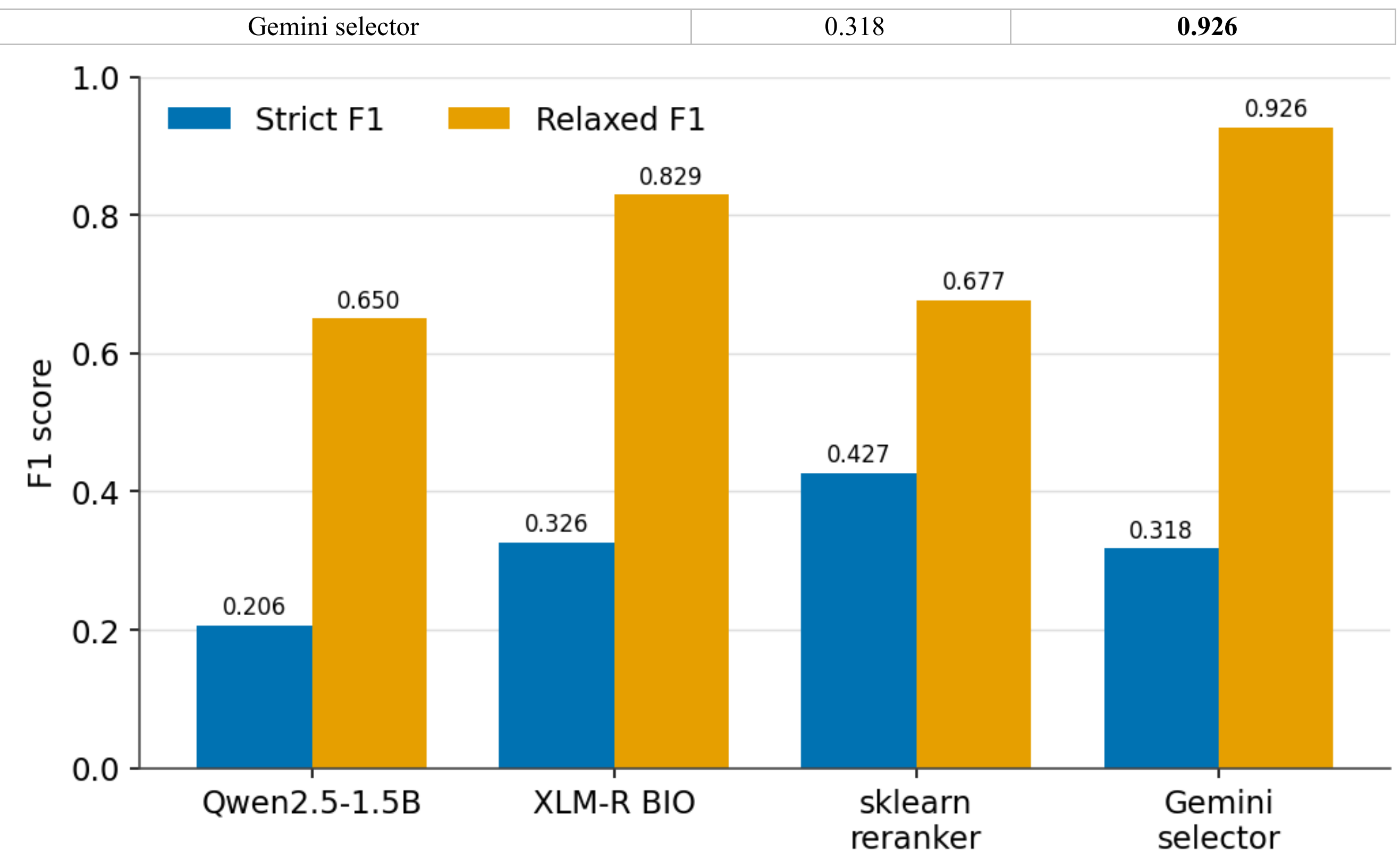

***Figure 1.*** *Clinical-decision extraction: strict versus relaxed F1 across models.*

### 3.3 Telemedicine and Patient-Portal Messaging

In telemedicine and portal messaging the patient writes asynchronously to the care team; processing is required because staff cannot read every incoming message in time, so models must triage, route, and prioritize them. Portal messaging is now studied both as a generation target, synthesizing realistic messages from a small de-identified seed[66], and as a triage and routing problem[67]. The six studies below span triage, severity scoring, distress classification, and home-care status monitoring.

#### *3.3.1 Administrative Triage of Patient-Portal Messages*

Patient-portal intake messages arrive short, informal, and error-strewn, and a busy practice cannot read each one in time. This study turns them into three administrative routing signals, framed as three supervised sub-tasks: an urgency level (green, yellow, or red), a multi-label set of risk factors, and a binary judgment of whether the message carries enough information to act on, leaving the decision itself with a human. Recent real portal-message data such as PMR-Bench[108] frames urgency as pairwise ranking rather than the multi-axis routing needed here, and portal messages remain protected health information, so no suitable corpus can be assembled.

The corpus is therefore entirely synthetic: a small local model, Qwen2.5-0.5B-Instruct, produces 3,000 message-and-label tasks, of which 2,799 parse cleanly, each message carrying its own structured labels. Three specialist DistilBERT heads, one per sub-task, are then compared against the same local model prompted zero- and few-shot (Table 11, Figure 2). The fine-tuned heads decisively outperform the prompted model, and their composition routes reliably, indicating that a small, on-premises generator can bootstrap a competent triage model in a controlled setting without any real data leaving the institution.

***Table 11.*** *Portal-message triage: per-head and pipeline results (synthetic test).*

| Component | Metric | Score |
|---|---|---|
| Urgency (DistilBERT) | accuracy / macro-F1 | 0.810 / 0.810 |
| Risk factors (DistilBERT) | micro-F1 | 0.879 |
| Insufficient-info (DistilBERT) | accuracy | 0.869 |
| Qwen (zero-shot, urgency) | macro-F1 | 0.295 |
| Composed pipeline (50 msgs) | urgency acc / risk exact-match | 0.76 / 0.92 |

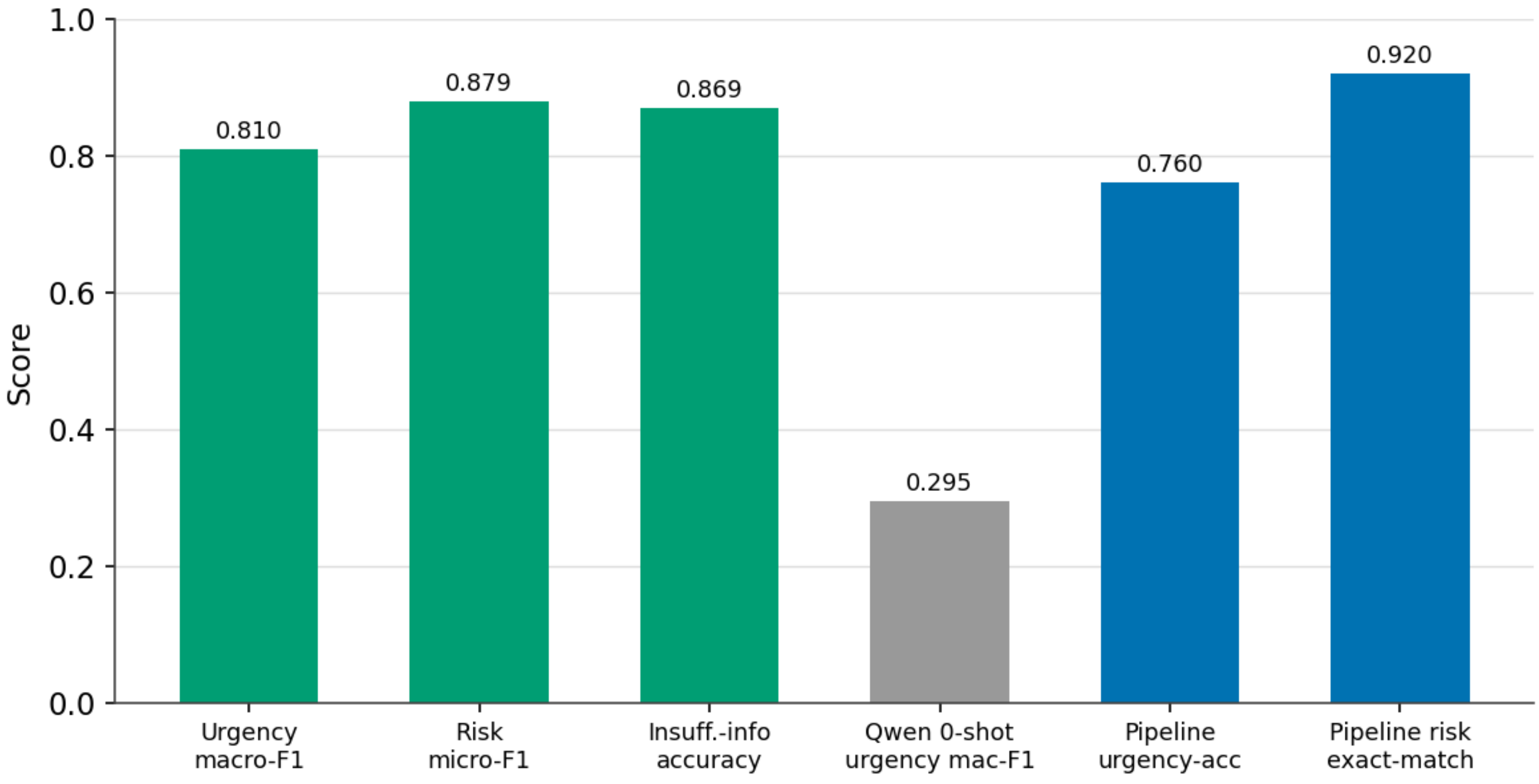

***Figure 2.*** *Portal-message triage: fine-tuned heads versus zero- and few-shot prompting.*

### 3.3.2 Severity Triage of Postpartum Messages

After a caesarean section, patients message their care team with concerns spanning routine soreness to hemorrhage, and safe after-care depends on sorting them quickly. This study frames the problem as four-class severity classification under the Manchester scheme, implemented as a two-stage cascade that first separates urgent from routine and then scores severity among the urgent cases. No labeled corpus of postpartum triage messages exists and the content is sensitive, so the data cannot be drawn from real records.

GPT-4o-mini, sampled at temperature 1.3 for diversity, therefore generates roughly 3,000 messages in which a writer profile and the target severity are fixed in the prompt, so the intended labels come from the source scenario, though their agreement with the generated message still requires validation. A BioBERT cascade and a Bi-LSTM cascade are compared against zero-shot GPT-4o-mini (Table 12, Figure 3). Notably, the zero-shot model matches the fine-tuned BioBERT cascade (0.98 versus 0.975 macro-F1), and both far exceed the Bi-LSTM (0.839), whose errors concentrate in the middle severities where the clinical distinctions are finest.

***Table 12.*** *Postpartum severity triage: four-class results (synthetic test, n=750).*

| Model | Accuracy | Macro-F1 |
|---|---|---|
| BioBERT cascade | 0.975 | 0.974 |
| GPT-4o-mini (zero-shot) | **0.980** | **0.980** |
| Bi-LSTM cascade | 0.836 | 0.839 |

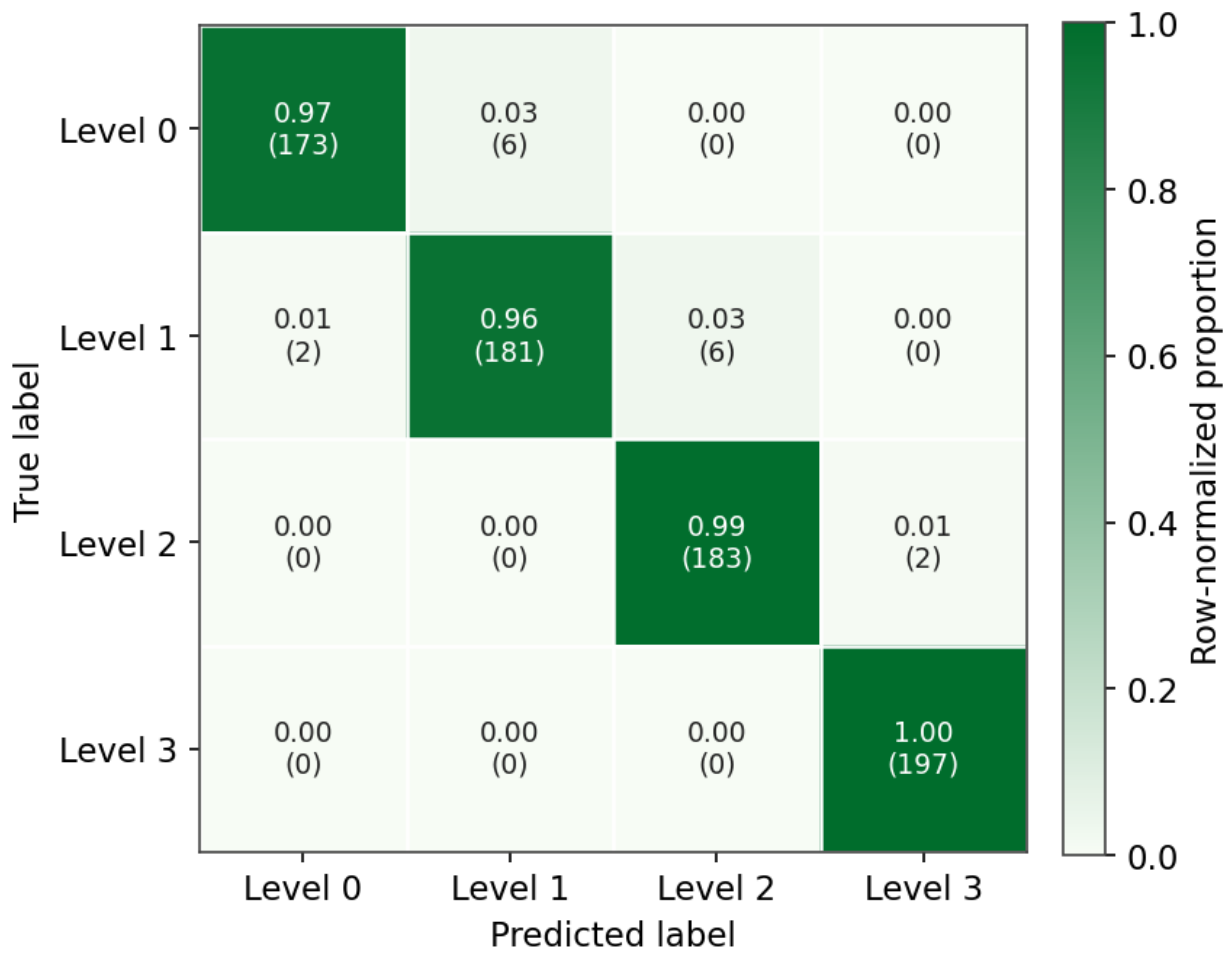

***Figure 3.*** *Postpartum severity triage: GPT-4o-mini zero-shot confusion matrix.*

### 3.3.3 Clinical-Priority Triage of Longitudinal Portal Messages

Beyond a single message, a patient's portal history must be sorted by clinical risk while holding a difficult balance between missed emergencies and alert fatigue. This study frames that as multi-class risk classification over longitudinal message sequences, with an explicit safety objective of never under-triaging an emergency. Existing portal datasets are single-message, and none provide longitudinal per-patient sequences with the multi-system priority criteria involved, so a real corpus is unavailable.

GPT-4o generates 3,000 longitudinal cases from structured clinical attributes and varied linguistic profiles, deliberately seeding "confidently wrong" edge cases to stress the classifier. Zero-shot LLMs, a fine-tuned DistilBERT, and a Bio_ClinicalBERT baseline are compared against a Bio_ClinicalBERT model wrapped in a conversational safety cascade (Table 13); the cascaded model attains near-perfect accuracy (0.997) with perfect recall on emergencies, far above the alternatives (0.34–0.42 for zero-shot LLMs, 0.828 for the plain encoder), which misclassify the high-risk class. The result shows how a safety-oriented layer trained on adversarial synthetic cases strengthens a classifier's handling of the high-risk class in this controlled evaluation.

***Table 13.*** *Clinical-priority triage of portal messages: accuracy (synthetic test).*

| Model | Accuracy |
|---|---|
| Zero-shot Qwen / Mistral-7B / BART-MNLI | 0.34–0.42 |
| DistilBERT (fine-tuned) | 0.817 |
| Bio_ClinicalBERT (baseline) | 0.828 |
| Bio_ClinicalBERT + safety cascade | **0.997** |

### 3.3.4 Classifying Psychosocial Distress in Oncology Messages

An oncology patient's message can carry more distress than its words admit, and support teams cannot read every one. This study infers, from a single message, both the dominant psychosocial response over seven categories and an ordinal distress level, a nominal and an ordinal classification task in parallel. Public cancer-forum sentiment sets exist, but they cover public posts rather than private secure messages and use emotion rather than psychosocial-response labels; as the project observes, there is no public labeled dataset for this framing, and emotional labels are subjective and expensive to annotate.

A local Gemma2-27B therefore generates 1,273 examples whose text never names the target label, a deliberate leakage control, after which a dual-model judge (GPT-4o-mini and Qwen2.5-32B) assigns a quality tier to each. A fine-tuned DistilBERT, a TF-IDF logistic-regression baseline, and zero-shot BART are compared (Table 14). DistilBERT leads on the ordinal distress task, reaching a linear-weighted $\kappa$ of 0.851, while the TF-IDF baseline edges ahead on the nominal response task; both far surpass zero-shot BART.

***Table 14.*** *Psychosocial-response and distress classification in oncology messages: macro-F1 by task (synthetic test).*

| Task | TF-IDF + LR | DistilBERT | BART zero-shot |
|---|---|---|---|
| Psychosocial response (7-way) | **0.856** | 0.834 | 0.732 |
| Distress level (3-level) | 0.805 | **0.864** | 0.468 |

### 3.3.5 Risk Classification of Medication Questions

Patients ask about their medications constantly, and a small fraction of those questions signal a genuine safety or interaction risk that should be escalated. This study frames the problem as binary classification, critical versus general, over medication questions. Adjacent resources such as MedicationQA carry no criticality label, and critical cases are intrinsically rare, so any real collection is severely imbalanced.

Two reviewers first label 5,000 real forum questions; GPT-4.1 then generates additional synthetic critical questions to rebalance the training set, a targeted use of generation as class-balancing augmentation that raises the GPT-4.1 classifier's own macro-F1 from 0.78 to 0.85. A classical SVM, GPT-4.1 in two modes, and fine-tuned BioBERT and BlueBERT are compared (Table 15); the fine-tuned encoders are strongest overall at 0.92 accuracy and 0.90 macro-F1. The held-out questions here are authentic patient-authored text and synthetic generation is confined to training-side rebalancing, which makes this the one study in the survey whose reported metric is measured on real communication rather than on held-out synthetic data; the classical SVM baseline (0.84 accuracy, 0.80 macro-F1) isolates that real-only signal. The study is reported in full as an analysis of predicting health crises from online medication inquiries[3].

***Table 15.*** *Critical-versus-general triage of medication questions.*

| Model | Accuracy | Macro-F1 |
|---|---|---|
| SVM (TF-IDF) | 0.84 | 0.80 |
| GPT-4.1 (classify only) | 0.87 | 0.78 |
| GPT-4.1 (generate + classify) | 0.89 | 0.85 |
| BioBERT | **0.92** | **0.90** |
| BlueBERT | **0.92** | **0.90** |

#### *3.3.6 Detecting Status Change in Home-Care Messages*

In home care no clinician is present, so the earliest signs of deterioration appear in the messages a patient sends and in their vital-sign trends. This study frames the problem as three-class status classification, no change, improvement, or deterioration, from free-text messages combined with vital-sign deltas, a text-plus-tabular fusion task. Home-telemonitoring datasets provide vital-sign streams but no paired patient messages or status-change labels, so the two modalities cannot be joined in real data.

The dataset is therefore generated synthetically, with model prompting over coupled symptom and vital patterns, rendered as free text plus deltas for heart rate, respiratory rate, temperature, and blood pressure. Text-only (TF-IDF, BERT), vitals-only (XGBoost, LightGBM), and late-fusion models are compared (Figure 4); fusing the two modalities wins, with a LightGBM model over TF-IDF text and vitals reaching 0.971 accuracy and 0.972 macro-F1, confirming that the message adds signal beyond the vitals alone.

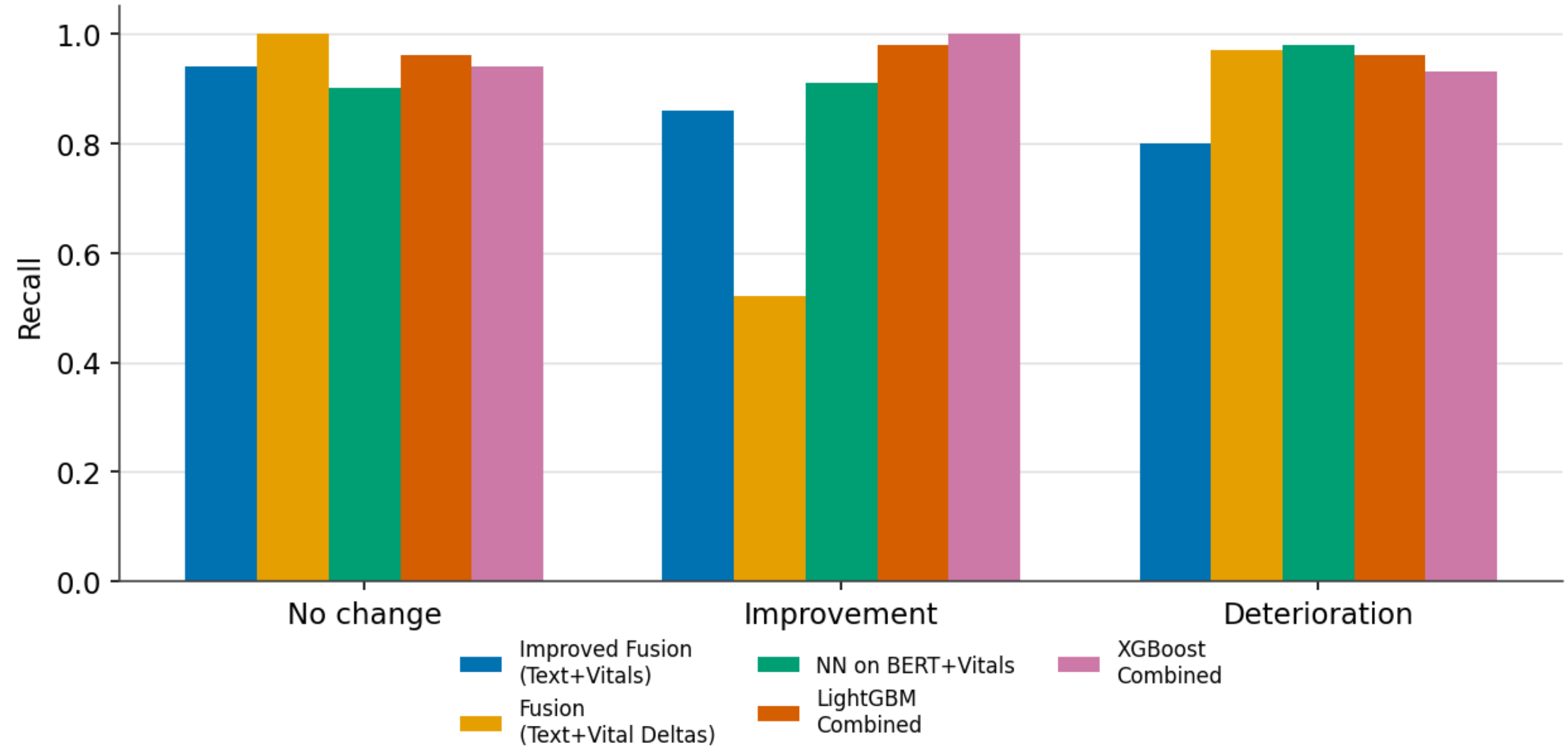

***Figure 4.*** *Home-care status detection: per-class recall of the top fusion models.*

### 3.4 Paramedic-to-Hospital, Dispatch, and Handoff

These are handovers between a paramedic, caller, or outgoing clinician and a hospital, dispatcher, or incoming team; processing matters because the communication is spoken, time-critical, and error-prone, and its structure must be recovered before information is lost. These channels are the least resourced and the most recently addressed: synthetic multi-party EMS dialogue[62], dispatch simulation[63], and LLM-generated handoff notes[65] have all appeared only lately, and dedicated synthetic SBAR corpora remain largely absent. The three studies below target EMS pre-arrival, field casualty radio, and SBAR handover.

#### 3.4.1 Routing EMS Pre-Arrival Reports

When an ambulance calls ahead, the emergency department must infer, from a hurried and noisy report, which care area to ready and which specialty to summon. This study frames both as classification, an eighteen-way specialty-consultation task alongside care-area routing, from EMS pre-arrival reports. EMS narrative NLP has been demonstrated only on private agency data, with no public corpus mapping reports to care-area and specialty labels, and real reports are short, incomplete, spoken over noise, and mangled by transcription.

GPT-4.1-mini generates 8,556 reports in four stylistic registers, from professional-and-complete to distracted handoff, over 2,139 MIMIC-IV-Ext source cases, while an audio-and-ASR pipeline injects realistic transcription noise. BioClinicalBERT and DistilBERT are trained with and without the noisy variants. The eighteen-way targets make this an intrinsically hard task, so absolute macro-F1 is low, but the informative result is that training on noisy synthetic reports improves robustness: removing the noisy variants lowers macro-F1 by 6.6 points on care area and 10.6 on specialty (Table 16, Figure 5). Deliberately degraded synthetic communication thus earns its value through realism as much as volume.

***Table 16.*** *EMS report routing: BioClinicalBERT macro-F1 (%) with and without noisy synthetic training.*

| Target | Train on clean + noisy | Train on clean only |
|---|---|---|
| ED care area | **29.2** | 22.6 |
| Specialty consultation | **38.3** | 27.7 |

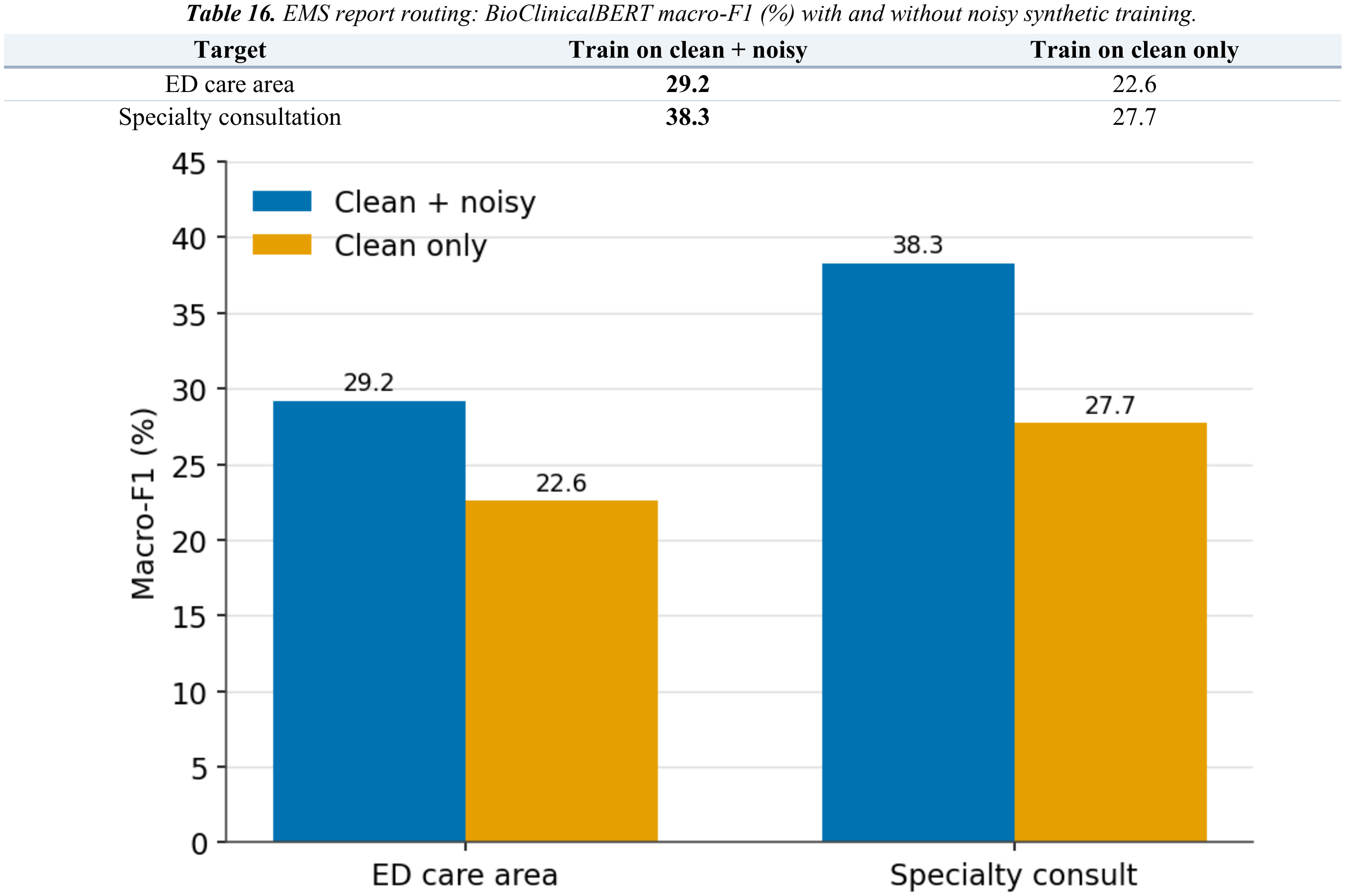

***Figure 5.*** *EMS report routing: macro-F1 across training setups; noisy synthetic data helps.*

#### 3.4.2 Reconstructing Casualty Records from Field Radio

On the battlefield only about half of casualty encounters are documented at all, and what reaches a hospital arrives as chaotic radio. This study reconstructs a structured twenty-field casualty record, an IDF Form 101, from that traffic, a multi-field slot-filling task over noisy Hebrew speech transcripts. Because such radio is inherently private, largely unrecorded, and effectively impossible to collect at scale, with no public dataset in this language, real training data does not exist.

GPT-4o, with fifty further samples from GPT-4o-mini, converts structured casualty profiles into 500 messy Hebrew transcripts, injecting static, communication cutouts, slang, and self-corrections at three reliability levels and emitting the ground-truth record alongside each transcript. An AlephBERT multi-head classifier is compared against GPT-4 in zero- and one-shot prompting and a QLoRA-fine-tuned Phi-3-mini (Table 17, Figure 6). The fine-tuned Hebrew encoder leads on exact-match (79.3%) and macro-F1 (0.802), while the small fine-tuned Phi-3 produces the fewest hallucinations (7), a property of real value when the output is a casualty record.

***Table 17.*** *Casualty-record reconstruction from field radio (synthetic test, n=100).*

| Model | Exact-match | Macro-F1 | Hallucinations |
|---|---|---|---|
| AlephBERT (multi-head) | **79.3%** | **0.802** | 16 |
| GPT-4 (zero-shot) | 68.6% | 0.698 | 24 |
| Phi-3-mini (QLoRA) | 63.1% | 0.632 | **7** |

***Figure 6.*** *Casualty-record reconstruction: accuracy, F1, and hallucination counts.*

### 3.4.3 Completeness Checking of SBAR Handovers

Shift handovers are among the most error-prone moments in care, and the SBAR format exists to make them complete. This study turns completeness itself into a supervised target, detecting missing or under-specified items in SBAR handover notes at both the item and the note level. The SBAR literature is confined to quality-improvement studies with no public annotated corpus, and handover notes are internal and sensitive, so labeled data must be created rather than collected.

OpenAI batch generation produces 5,000 synthetic SBAR notes over a requiredness-and-completeness schema with both per-item and note-level labels. A range of models is compared, from majority and style baselines and TF-IDF logistic regression to text-only and context-aware BioClinicalBERT, Qwen prompting, and a hierarchical extension that predicts requiredness before completeness (Table 18, Figure 7). Fine-tuned encoders outperform prompting by a wide margin (0.745 versus 0.408 class macro-F1), and the requiredness-first structure lifts class macro-F1 further to 0.791, offering a reusable recipe for handoff auditing.

***Table 18.*** *SBAR handover completeness: note-level results (synthetic test).*

| Model | Accuracy | Class macro-F1 | Safety-missing recall |
|---|---|---|---|
| TF-IDF logistic regression | 0.725 | 0.711 | 0.742 |
| BioClinicalBERT (text-only) | 0.758 | 0.745 | 0.756 |
| Qwen (zero-shot) | 0.489 | 0.408 | – |
| Hierarchical (requiredness-first) | – | **0.791** | – |

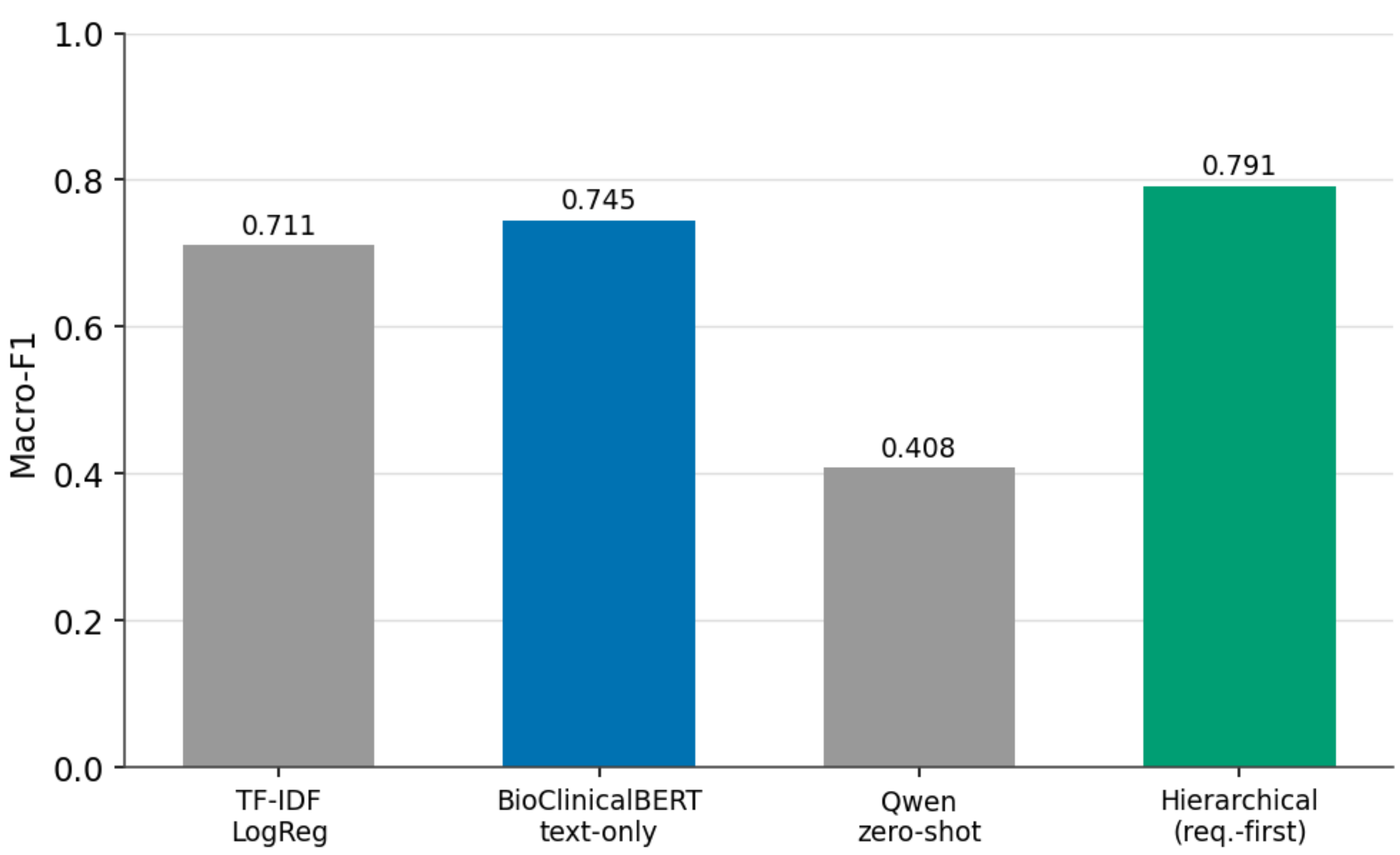

***Figure 7.*** *SBAR handover completeness: class macro-F1 across models.*

## 4 Limitations and Caveats of Synthetic Data

Synthetic clinical communication is useful precisely because it is not real, and that same fact is the source of its limitations. This section states them plainly, so that the enthusiasm of Section 3 is read with appropriate caution. We separate technical limitations of the data itself (4.1), practical and ethical caveats around its use (4.2), and a reminder of what synthetic data cannot replace (4.3).

### 4.1 Technical Limitations

The technical obstacles to synthetic clinical communication are, at root, obstacles of fidelity. The most consequential is clinical hallucination[82]: a generator can produce fluent text that is medically wrong, inventing a contraindication, an implausible vital sign, or a treatment that does not follow from the case. Because such errors are locally coherent, they slip past surface fluency checks and can silently corrupt the very labels a downstream model learns from, which makes factual control, through grounding, constrained decoding, synthetic edit feedback[76], and post-hoc verification, the central design problem rather than an afterthought. A second difficulty is the long tail: the rare conditions and dangerous edge cases that make synthetic data attractive are also the hardest to render faithfully, because the generator has seen fewest examples of them and is most prone to stereotype or omission. Third, even when individual messages are realistic, their aggregate may exhibit distribution shift relative to genuine communication, with different lexical statistics, error patterns, or class balance, so that a model trained on synthetic data underperforms on real text in ways that held-out synthetic evaluation cannot detect. Longitudinal consistency compounds the problem, since a patient's story must remain coherent across a conversation, a handoff, and a discharge, yet independently sampled turns readily contradict one another. Beneath all of these lies a measurement gap: evaluation metrics that reliably predict downstream utility are absent, and current fidelity and diversity measures are demonstrably imperfect[80,81]. This is why the train-on-synthetic, test-on-real protocol, still uncommon in the studies surveyed here, remains the field's most trustworthy yardstick.

### 4.2 Practical and Ethical Caveats

Beyond the technical lie institutional and ethical hurdles. Regulatory acceptance is immature: it is not yet settled what validation a synthetic corpus must pass before a model trained on it may be used in care, nor which bodies should certify that validation. The ethics cut in two directions. Synthetic data is often justified on privacy grounds, yet the assumption that it is automatically safe is unfounded, because synthetic notes that match real utility can inherit real privacy risk[31], a risk that membership-inference audits now make measurable[77] and that differentially private generators address only imperfectly on specialist clinical text[78]; and there is a disclosure obligation, since text that reads as a genuine patient encounter should be labeled as synthetic wherever it circulates. Sharing and licensing remain unsettled, particularly when a generator was conditioned on restricted real data, which can quietly propagate governance constraints into ostensibly open outputs. Finally, reproducibility is fragile: results depend on the exact generator, prompt, sampling temperature, and filtering pipeline, few of which are reported in enough detail to reconstruct, so studies that appear comparable may not be comparable at all.

### 4.3 Boundaries of the Method

Some limitations are boundaries of the method rather than defects to be engineered away. Synthetic communication reflects the knowledge, and the blind spots, of the model that produces it: it faithfully reproduces the distributions the generator has learned, while genuinely novel presentations, idiosyncratic phrasing, and culturally specific idiom, often the hardest cases for clinical NLP, still call for authentic data. Validation carries the same qualification, since a model assessed only on synthetic communication has been measured against a world its own family generated; authentic-data evaluation therefore remains the decisive test. Synthetic data is best understood as a scaffold: indispensable for building and stress-testing systems, and complemented by the real communication and real patients that ultimately establish readiness for care.

## 5 Research Frontiers and Open Problems

The clearest opportunities cluster into a few themes. The first is **interactive simulation**. Moving from static corpora toward multi-agent simulation, in which patient and clinician agents generate consultations on demand and double as graded evaluation environments, fits the sequential, goal-directed nature of clinical talk. Coevolving standardized-patient agents[60], safety-focused simulation-and-evaluation frameworks[98], and multimodal clinical-agent benchmarks[99] point toward environments that fold generation and evaluation into one renewable loop, extending toward personalized synthetic patients[10,54,55,56].

A second theme is **spoken and multi-party communication**. The EMS, dispatch, and verbal-handoff channels are spoken first and rich in protected information, yet text-only synthesis discards the acoustic-error distribution that dominates real deployment. Joint synthetic pipelines that pair generated audio with transcripts and inject realistic recognition noise are a clear open space that would let transcription-robust clinical NLP be trained without recording real calls.

A third is **privacy, fidelity, and utility, co-measured**. Membership-inference and canary auditing now quantify leakage on the exact synthetic text behind a utility claim[77], fidelity and diversity metrics are under active and critical revision[80,81], and differentially private generators still contend with specialist clinical distributions[78]. The open problem is to co-optimize and co-report privacy, clinical faithfulness, and downstream utility in a single pass on one generator, rather than in separate studies.

A fourth is **controllable, causal, and continually current generation**. Explicit control over uncertainty, omission, code-switching, and recognition noise, together with counterfactual variants of the same encounter, would let downstream models be probed for what they actually rely on; surveys of conversational data generation catalog the available levers[102]. Continual, self-refining generation could keep synthetic communication current as guidelines and language drift, provided recursive self-training is guarded against model collapse[101].

Underlying all of these are **foundations, benchmarks, and human oversight**. Whether synthetic clinical dialogue obeys predictable scaling laws, and where returns plateau, is an open and decision-critical question[100]; shared, communication-specific fidelity benchmarks, including rubric-scored and OSCE-style stations[104], would let scaling laws, privacy audits, and agent evaluations be compared across groups. Removing generator fingerprints so that models learn clinical content rather than stylistic tics[103], preference optimization toward clinician-judged realism, and automated safety auditing that treats omission as seriously as fabrication[82] complete an agenda in which clinician oversight remains central.

## 6 Conclusions

LLM-generated synthetic clinical communication has matured from a privacy-preserving stand-in for real records into a versatile resource for building, evaluating, and benchmarking clinical NLP systems. This survey organized the field around the communication channel, from patient–clinician conversations and doctor-to-patient instructions to telemedicine and portal messaging and the under-served paramedic, dispatch, and handoff channels, and around the downstream tasks each supports, grounding that organization in thirteen application studies that generate synthetic communication and train models on it.

Read together, those studies surface a small set of reusable design patterns (Table 19) that recur across channels and tasks rather than one-off results, and that answer RQ7 directly.

***Table 19.*** *Reusable design patterns recurring across the thirteen application studies (RQ7).*

| Pattern | What it does | Studies |
|---|---|---|
| Fine-tuned encoders over zero-shot LLMs | A small encoder fine-tuned on synthetic data outperforms the evaluated zero-shot LLM baselines on the downstream task | §3.1.1, §3.1.3, §3.3.1, §3.3.4, §3.3.5, §3.4.1, §3.4.3 |
| Deliberate degradation for robustness | Injecting noise, recognition errors, or omissions into the synthetic text trains models that hold up on degraded real-world input | §3.1.1, §3.4.1, §3.4.3 |
| Labels fixed at generation, audited by a judge | Gold labels are emitted with the text and an independent LLM judge audits quality, giving both control and label reliability | §3.1.2, §3.2.1, §3.3.4 |

| Synthetic augmentation of rare classes | Generation up-samples the rare, high-stakes classes that real collections underrepresent, correcting severe imbalance | §3.3.2, §3.3.5 |
|---|---|---|
| Local, on-premises generation for privacy | A local generator bootstraps a usable dataset without sending any clinical content to a hosted model | §3.3.1 |

They come, however, with an unmistakable caveat. Almost every study evaluates on held-out synthetic data. The clearest exception is the medication-question study (Section 3.3.5), whose classifier is tested on authentic patient-authored questions with synthetic generation confined to rebalancing the rare class in training; two further studies ground generation in real clinical records, tracing decision extraction to MedDec and MIMIC-III annotations (Section 3.2.1) and EMS routing to MIMIC-IV cases (Section 3.4.1). Even so, the decisive evidence, strong performance when a model is trained end-to-end on synthetic communication and tested on real communication, is still largely outstanding. Closing that gap, alongside progress on hallucination control, distributional fidelity, longitudinal consistency, and validation standards that regulators will accept, is the work that will decide whether synthetic clinical communication becomes durable infrastructure or remains a convenience. As generation quality, evaluation rigor, and privacy guarantees advance together, and as agent-based patient simulation turns static corpora into interactive environments, it is well positioned to become the former: a reusable resource that lets the field learn from the conversations central to care while protecting the people who have them.

## Data and Code Availability

The datasets, generation scripts, and model configurations for all thirteen application studies in Section 3 are shared publicly in the paper's repository under CaseStudies/ and archived at Zenodo (doi:10.5281/zenodo.21820227), organized as one directory per study, each with a README documenting its motivation, data-generation protocol, data files, models, and results. The regenerated result figures are distributed with the paper's source. Where a study reused a restricted real dataset for evaluation, only the synthetic artifacts and code are redistributed, in line with the original data licenses.

## Acknowledgments

We warmly thank the students of the LLMs in Healthcare course in the Digital Healthcare department at the Holon Institute of Technology (HIT), across the 2024–2025 and 2025–2026 academic years. Their course projects motivated this survey and supplied the application studies in Section 3, showing hands-on that LLM-generated synthetic clinical communication can drive real clinical NLP systems. We thank the whole cohort collectively for their creativity and rigor.